# Teaching in adverse scenes: a statistically feedback-driven threshold and mask adjustment teacher-student framework for object detection in UAV images under adverse scenes

Hongyu Chen [a, b], Jiping Liu [a, b] *, Yong Wang [b], Jun Zhu [a], Dejun Feng [a], and Yakun Xie [a]

[a] Faculty of Geosciences and Environmental Engineering, Southwest Jiaotong University, Chengdu, China

[b] Research Center of Geospatial Big Data Application, Chinese Academy of Surveying and Mapping, Beijing, China

* Correspondence: liujp@casm.ac.cn (Jiping Liu)

**Abstract:** Unmanned Aerial Vehicles (UAVs) have become a key platform for aerial object detection, but their performance in real-world scenarios is often severely impacted by adverse environmental conditions, such as fog and haze. Achieving robust UAV object detection under these challenging conditions is crucial for enhancing the all-weather situational awareness capabilities of UAVs. This is especially critical in key application scenarios, such as rapid disaster response and information interpretation, which demand reliable visual perception around the clock. Unsupervised Domain Adaptation (UDA) has shown promise in effectively alleviating the performance degradation caused by domain gaps between source and target domains, and it can potentially be generalized to UAV object detection in adverse scenes. However, existing UDA studies are based on natural images or clear UAV imagery, and research focused on UAV imagery in adverse conditions is still in its infancy. Moreover, due to the unique perspective of UAVs and the interference from adverse conditions, these methods often fail to accurately align features and are influenced by limited or noisy pseudo-labels. To address this, we propose the first benchmark for UAV object detection in adverse scenes, the Statistical Feedback-Driven Threshold and Mask Adjustment Teacher-Student Framework (SF-TMAT). Specifically, SF-TMAT introduces a design called Dynamic Step Feedback Mask Adjustment Autoencoder (DSFMA), which dynamically adjusts the mask ratio and reconstructs feature maps by integrating training progress and loss feedback. This approach dynamically adjusts the learning focus at different training stages to meet the model's needs for learning features at varying levels of granularity. Additionally, we propose a unique Variance Feedback Smoothing Threshold (VFST) strategy, which statistically computes the mean confidence of each class and dynamically adjusts the selection threshold by incorporating a variance penalty term. This strategy improves the quality of pseudo-labels and uncovers potentially valid labels, thus mitigating domain bias. Extensive experiments demonstrate the superiority and generalization capability of the proposed SF-TMAT in UAV object detection under adverse scene conditions. The Code is released at https://github.com/ChenHuyoo .

# 1 Introduction

In fields such as autonomous driving, video surveillance, intelligent transportation, and the increasingly vital area of disaster emergency response, object detection technology has become a core tool for understanding and perceiving the surrounding environment (Jeon et al., 2024; Zhong et al., 2024; Gupta et al., 2024; Feng et al., 2023). Owing to their cost-effectiveness and versatility, object detection based on images from Unmanned Aerial Vehicle (UAV) perspectives has also emerged as a crucial research area (Li et al., 2025a; Chen et al., 2025a). However, in outdoor environments, UAV-perspective imagery is often affected by adverse weather conditions. This challenge of environmental adaptability is particularly prominent in critical applications, such as disaster information interpretation, which require all-weather, rapid response capabilities, because adverse weather is often an unavoidable component of such scenarios. Adverse weather not only leads to image quality degradation but also introduces issues of inter-domain discrepancy (domain shift). Furthermore, the scale variations and non-uniform feature distributions inherent to the UAV perspective further exacerbate the performance decline of detection models.

In recent years, with the rapid development of deep learning, numerous object detection methods have emerged. These methods have demonstrated remarkable success on high-quality and clear images (Chen et al., 2025b; Wang et al., 2025; Wang et al., 2024a). However, they often assume that the feature distributions of the training and testing data are consistent, which does not always hold in real-world outdoor scenarios, particularly under foggy weather conditions. Foggy weather is one of the most frequently occurring adverse conditions in practical applications, where the presence of fog blurs or obscures object features, leading to a decline in detection accuracy. To address this issue, researchers have proposed two primary approaches: one involves embedding a denoising module before the object detection model (Li et al., 2023a; Hu et al., 2024), while the other incorporates prior knowledge into the training process (Zhong et al., 2024). However, these methods may degrade detector performance and often require a complex design process. Moreover, existing studies primarily focus on adverse conditions in natural images (Zhang et al., 2022; Liu et al., 2022b), while research on the impact of adverse weather conditions in remote sensing images remains largely unexplored.

In remote sensing scenarios, object detection faces unique challenges due to significant variations in UAV flight altitude, viewing angles, and scene coverage (Feng et al., 2024a; Du et al., 2023). While numerous studies have achieved remarkable progress in UAV image-based object detection (Li et al., 2025b; Wang et al., 2025; Lin et al., 2025; Li et al., 2023b), including weak object detection to enhance the feature response of targets of interest (Han et al., 2021; Han et al., 2022), small object detection to improve discriminative feature learning for tiny objects (Zhang et al., 2025; Yan et al., 2025), and limited-supervision object detection under insufficient annotation conditions (Wang et al., 2025; Wang et al., 2023), object detection on UAV images under adverse conditions remains underexplored. Thanks to the HazyDet dataset proposed by Feng et al. (2024a), object detection in foggy UAV imagery has gained an initial foothold. Alongside the dataset, they introduced DeCoDet, the first detection network tailored for UAV imagery in foggy conditions. However, most object detection methods in remote sensing scenarios rely heavily on large-scale annotated datasets for supervised learning to achieve impressive performance. Obtaining accurate large-scale labeled bounding boxes is inherently challenging. Moreover, when subject to domain shifts, pretraining data and target data exhibit significant distribution discrepancies, as illustrated in Figure 1(a), leading to a notable decline in detection performance.

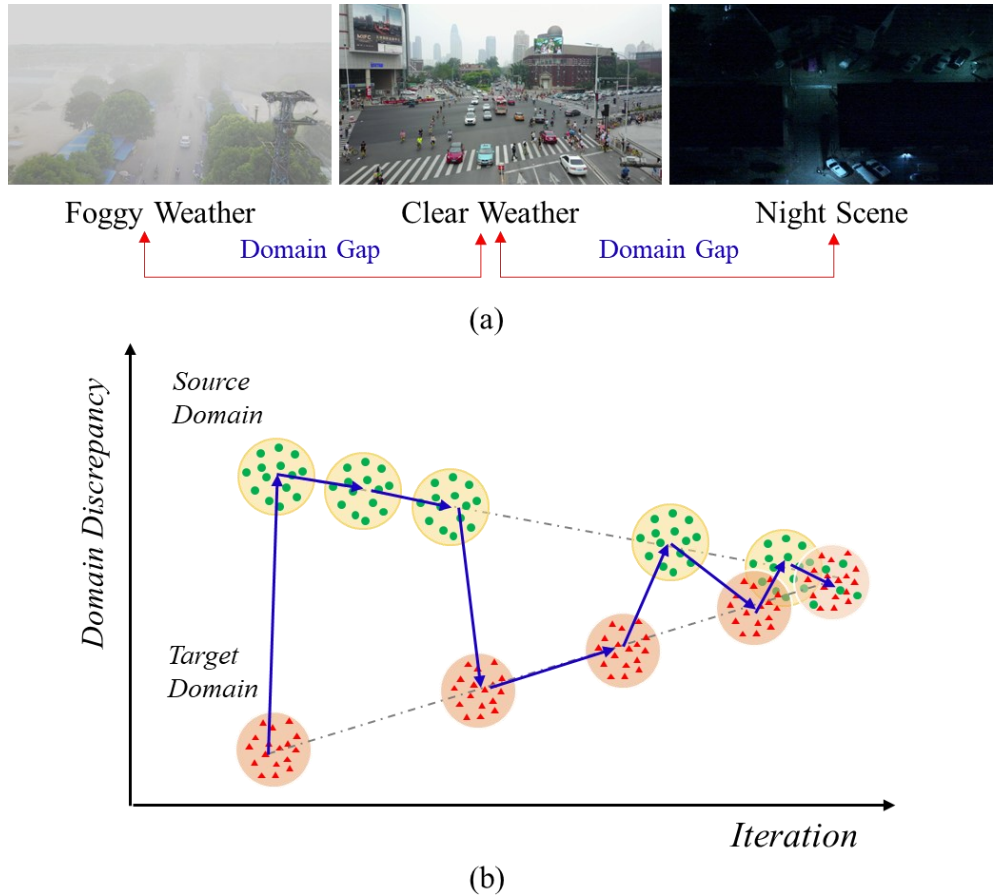

**Fig. 1.** Proof of domain gap and example of domain alignment. (a) Proof of domain gap under different environmental conditions; (b) Example of the model gradually achieving domain alignment as training iterations progress.

To address the issue of domain shift without additional annotations, numerous unsupervised domain adaptation (UDA) methods have been proposed. These approaches facilitate domain alignment by jointly training on labeled source domain data and unlabeled target domain data, as illustrated in Figure 1(b). Research on UDA for object detection primarily leverages techniques such as adversarial alignment, image-to-image translation, graph neural networks (GNNs), and mean teacher training to enhance detector performance. However, most of these methods are designed for natural images (Jeon et al., 2024; Guo et al., 2024). In the remote sensing domain, UDA research has mainly focused on image classification and segmentation tasks, with relatively few studies addressing object detection tasks (Ma et al., 2024; Han et al., 2024; Cao et al., 2024; Feng et al., 2024b). Furthermore, due to the unique perspective of UAV imagery and the complexities introduced by adverse weather conditions, research on object detection in UAV imagery under adverse conditions remains largely unexplored. We observe that UAV imagery is inherently different from natural images. First, UAV images are typically captured from higher altitudes, leading to greater scale variations in objects. Second, UAV imagery often involves more complex backgrounds, where objects exhibit irregular spatial distributions. These factors make it challenging to directly transfer object detection algorithms developed for natural images to UAV imagery.

Among the aforementioned UDA methods, the Teacher-Student framework, wherein pseudo-labels generated by a teacher model guide the learning of a student model, has demonstrated exceptional performance in object detection tasks (He et al., 2022a; Chen et al., 2022). Notably, the MRT approach (Zhao et al., 2023) has achieved significant progress in domain adaptive detection for natural images. However, for UAV scenarios under adverse conditions, current research has yet to effectively apply this framework. Therefore, with respect to object detection in UAV imagery under adverse conditions and the Teacher-Student framework, we highlight three primary shortcomings that remain in existing research.

(1) Methodological Limitations: Current approaches heavily rely on annotated data and lack a strategy capable of achieving significant performance improvements without additional annotations.

(2) Limitations of the Teacher-Student Framework: Within existing Teacher-Student frameworks, including MRT, although MRT integrates Masked Autoencoders (MAE) into its training process, it utilizes a fixed and often high masking ratio. This rigid approach neglects the varying learning requirements for features at different granularities specific to UAV scenarios, as well as the potential need for different reconstruction difficulties due to domain shifts. A fixed high masking ratio risks losing crucial information, whereas a fixed low ratio might fail to pose an adequate challenge for effective self-supervised learning.

(3) Issues in Pseudo-Label Generation: Current methods typically utilize a fixed confidence threshold for filtering pseudo-labels. This approach may struggle to effectively handle the pronounced fluctuations in prediction confidence and significant inter-class variations inherent in UAV scenarios, which arise from factors such as complex backgrounds and diverse object scales. Furthermore, it disregards differences in sample distribution among classes and the dynamic nature of the training process. Such a fixed thresholding practice tends to introduce a considerable number of noisy pseudo-labels, particularly in later training phases when confidence scores generally rise, causing the model to overfit erroneous information.

In this paper, we propose an innovative UDA-based teacher-student framework for object detection in UAV imagery under adverse conditions. This framework effectively addresses domain shift issues without requiring additional annotated data while achieving superior performance. Addressing the limitation of existing methods where a fixed masking ratio struggles to adapt to the variable object scales and scene complexities inherent in UAV imagery, thereby hindering feature learning effectiveness, our SF-TMAT framework introduces a feedback-driven dynamic mask adjustment mechanism. This mechanism adaptively adjusts the reconstruction difficulty based on the actual information density and complexity of the input image, compelling the model to learn feature representations more robust to scale variations and background interference. Concurrently, to overcome the deficiency of relatively simple thresholding strategies that fail to cope with the drastic fluctuations in prediction confidence under adverse conditions, leading to poor pseudo-label quality, we employ a statistical variance-aware threshold smoothing strategy. This strategy not only considers the mean confidence but also utilizes variance to assess prediction stability, thereby enabling the intelligent preservation of true positive targets under adverse conditions that may have low confidence but also low variance, while effectively filtering out noisy pseudo-labels with high confidence fluctuation. Through these two key adaptive improvements, SF-TMAT can more effectively tackle the challenges posed by adverse UAV environments. Our main contributions are summarized as follows:

(1) We introduce a statistically feedback-driven threshold and mask adjustment teacher-student framework (SF-TMAT) for object detection in UAV images under adverse conditions, based on the Deformable DETR detector. To the best of our knowledge, this is the first study to explore UDA-based methods for object detection in UAV imagery under adverse conditions. Experimental results demonstrate that SF-TMAT effectively handles object detection challenges in such scenarios, achieving superior accuracy and robustness compared to state-of-the-art UDA methods, and even outperforming some supervised approaches.

(2) We propose a Dynamic Step Feedback Mask Adjustment (DSFMA) strategy for dynamically generating masks of varying complexity. Unlike the conventional approach of pretraining with Masked Autoencoders (MAE) followed by fine-tuning, DSFMA introduces a step size adjustment mechanism based on the Sigmoid function. By incorporating training progress and loss feedback, it dynamically adjusts the mask ratio, enhancing the model's adaptability across different training stages and improving its ability to learn domain-invariant features.

(3) To select more reliable pseudo-labels and dynamically adapt to the distribution characteristics of pseudo-labels, the Variance Feedback Smoothing Threshold (VFST) strategy is applied to the framework. This strategy dynamically adjusts the filtering threshold by combining the mean confidence and variance information. Specifically, VFST balances the threshold using a temporal dynamic smoothing coefficient and current feedback, while introducing a variance penalty term to suppress noisy pseudo-labels, thereby improving the quality of pseudo-labels during the teaching process.

## 2 Related work

### *2.1 Object detection in adverse scene condition*

To better meet the demands of real-world applications, many methods aim to develop a well-trained detector capable of operating in adverse environments. The most common approach is to train the detector using annotated datasets collected under adverse conditions. However, obtaining a large-scale, annotated dataset for adverse environments in real-world scenarios is highly challenging. As a result, some researchers synthesize weather effects onto clear images and train detectors using supervised learning. These approaches can be categorized into two paradigms: separate optimization (Li et al., 2023) and joint optimization (Qin et al., 2022; Appiah et al., 2024; Zhong et al., 2024). The separate optimization paradigm attaches a restoration network before the detection model and trains them independently——first restoring adverse weather images to clear images, followed by object detection. In contrast, the joint optimization paradigm integrates image restoration and object detection within a unified framework. For example, Li et al. (2023) designed a dual-branch structure with an attention fusion module that connects the dehazing module and detection module in an end-to-end manner. Similarly, Zhong et al. (2024) proposed a prior knowledge-guided network for end-to-end object detection in foggy conditions.

However, the separate optimization paradigm often fails to significantly improve object detection accuracy and may even lead to the loss of high-frequency details during the restoration stage. This issue becomes more pronounced in UAV imagery, where objects appear at varying scales. Although the joint optimization paradigm alleviates this problem to some extent, it still heavily relies on synthetic datasets and incurs higher computational costs. Moreover, in the UAV context, obtaining paired data from the source domain and the degraded domain is often impractical.

## *2.2 Object detection via domain adaptation*

Unsupervised Domain Adaptation (UDA) enables the direct transfer of object detectors trained on clear images to target domains with adverse weather conditions. This approach effectively mitigates domain discrepancies in object detection without requiring paired data from the target domain. Chen et al. (2018) pioneered the first UDA-based object detector by incorporating adversarial training into Faster R-CNN. Following this, numerous researchers explored adversarial feature alignment from various perspectives. With the rise of transformer-based architectures, some scholars specifically designed transformer-driven UDA object detectors for feature alignment. Subsequently, techniques such as image transformation, graph reasoning, and pseudo-label self-training have been progressively introduced into UDA-based object detection. Among these approaches, the teacher-student framework has gained prominence in object detection. For example, Belal et al. (2024) employed contrastive loss learning to align features of the same category, while Wang et al. (2024b) leveraged knowledge distillation to significantly enhance the detection capability of the student model. As UDA-based object detection has achieved remarkable success in natural images, its application has gradually expanded to remote sensing. For instance, Biswas et al. (2024) proposed a support-guided debiased contrastive learning method to mitigate bias caused by false negative samples, while Han et al. (2024) introduced the first teacher-student framework for object detection in remote sensing, built upon the DETR detector.

However, in the domain of natural images, existing teacher-student frameworks face significant challenges due to the presence of low-quality pseudo-labels. In the remote sensing domain, apart from the pioneering work of Feng et al. (2024a) on object detection under adverse conditions, all other methods have been developed based on clear images. Notably, two key factors hinder the direct transfer of object detection methods from natural images to remote sensing images. First, remote sensing imagery possesses unique characteristics that fundamentally differ from those of natural images, making conventional detection methods less effective. Second, while Feng et al. (2024a) significantly improved object detection performance in foggy conditions by incorporating depth information and supervised learning, their approach relies on auxiliary information and accurate target annotations. Both of which are often unavailable in real-world scenarios at the time of deployment. In contrast to existing approaches, the method proposed in this paper is fundamentally different. Our approach achieves significant improvements in object detection accuracy under adverse weather conditions without requiring additional target annotations for degraded images. This helps mitigate the severe performance degradation commonly observed in existing methods when exposed to adverse environmental conditions.

## 3 Proposed method

### 3.1 Network Overview

Figure 2 illustrates the statistically feedback-driven threshold and mask adjustment teacher-student framework (SF-TMAT) proposed in this paper, which is based on the Deformable DETR detector. Compared to traditional detectors, selecting DETR as the baseline detector for UAV scenarios offers multiple advantages. DETR's inherent global relationship modeling capability facilitates better understanding and localization of the sparse objects and complex backgrounds commonly found in UAV images. Moreover, DETR's simplified end-to-end pipeline eliminates hand-crafted components like anchor boxes and Non-Maximum Suppression (NMS), which simplifies the design of cross-domain feature alignment required to address the domain shifts frequently encountered in UAV imagery. This framework consists of three key components: Dynamic Step Feedback Mask Adjustment Autoencoder (DSFMA), Variance Feedback Smoothing Threshold Strategy (VFST), and Selective Retraining Strategy (SRS). DSFMA adaptively updates the step size according to the training progress, enabling dynamic adjustment of the masking ratio to enhance the model's adaptability (Section 3.2). VFST incorporates multi-source feedback information to dynamically update the class-specific thresholds during the teaching phase, thereby improving pseudo-label quality (Section 3.3). SRS employs a selective retraining mechanism to help the student model escape local optima (Section 3.4). The following subsections provide a detailed explanation of DSFMA and VFST, while SRS will be discussed in Section 4.2.

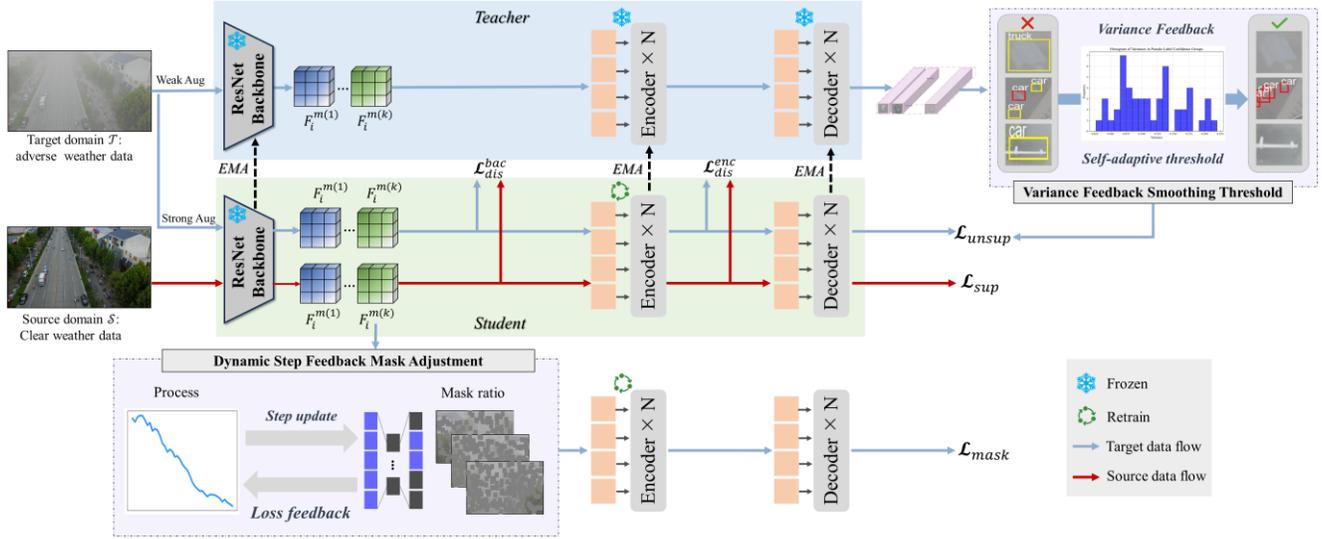

**Fig. 2.** The overall architecture of SF-TMAT.

## 3.2 Dynamic Step Feedback Mask Adjustment Autoencoder

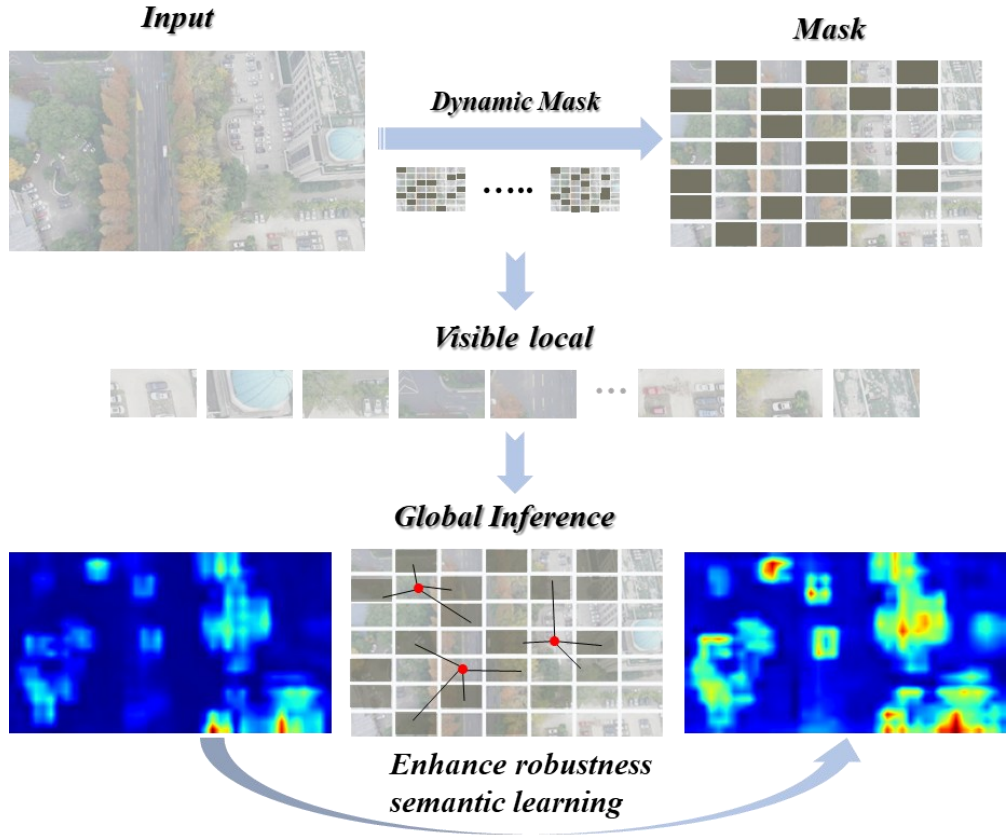

**Fig.3.** The overall process of DSFMA.

To better capture target domain features and enable the model to gain a deeper understanding of them, we design a personalized Masked Autoencoder (MAE) with dynamic mask ratio adjustment, as shown in Figure 3, with the pseudocode provided in Figure 4. In this approach, multi-scale feature maps of the target image are masked at varying ratios based on the training progress and then fed into the student model. Subsequently, the masked features are reconstructed using an auxiliary decoder. This self-supervised training paradigm compels the model to infer global semantics from visible local semantics, enabling it to learn more robust feature representations. The following section provides a detailed explanation of the DSFMA operation process.

**Algorithm 1** Dynamic Step Feedback Mask Adjustment
---
**Require:** Initial mask ratio ($\mu_t$), loss history ($\mathcal{L}_{\text{history}}$), current loss ($\mathcal{L}_{\text{current}}$), step size bounds ($\eta_{\min}, \eta_{\max}$), sigmoid parameters ($k$, midpoint), total epochs ($T_{\max}$), current epoch ($T_i$)
**Ensure:** Updated mask ratio $\mu_t$
1: Initialize: $\mu_t \leftarrow \mu_0$, $\eta \leftarrow \eta_{\max}$
2: **for** $i = 1$ to $T_{\max}$ **do**
3:    $x \leftarrow T_i / T_{\max}$
4:    $\sigma \leftarrow 1/(1 + \exp(-k \cdot (x - \text{midpoint})))$
5:    $\eta \leftarrow \eta_{\min} + (\eta_{\max} - \eta_{\min}) \cdot (1 - \sigma)$
6:    **if** $|\mathcal{L}_{\text{history}}| \geq 3$ **then**
7:       $\mathcal{L}_{\text{mean}} \leftarrow \text{average}(\mathcal{L}_{\text{history}}[-3:])$
8:    **else**
9:       $\mathcal{L}_{\text{mean}} \leftarrow \mathcal{L}_{\text{current}}$
10:   **end if**
11:   **if** $\mathcal{L}_{\text{current}} \geq \mathcal{L}_{\text{mean}}$ **then**
12:      $\mu_t \leftarrow \mu_t - \eta$
13:   **else**
14:      $\mu_t \leftarrow \mu_t + \eta$
15:   **end if**
16: **end for**
17: **return** $\mu_t$

**Fig.4.** Detailed pseudocode of DSFMA.

### *3.2.1 Feature Dynamic Masking*

Deformable DETR receives multi-scale feature maps from the backbone, denoted as $\{F_i \in \mathbb{R}^{C_i \times H_i \times W_i}\}_{i=1}^{K}$, where K represents the number of feature maps. Each feature map is then randomly masked with a binary mask $\{m_i \in \{0,1\}^{H_i \times W_i}\}_{i=1}^{K}$. Unlike traditional fixed masking ratio strategies (He et al., 2022b), the masking ratio $\mu_t$ in DSFMA is controlled with finer granularity. This adjustment relies on two core techniques: Sigmoid-based dynamic step adjustment and Loss-driven mask ratio update.

(1) Sigmoid-based dynamic step adjustment

The core idea of this adjustment mechanism is to dynamically regulate the masking ratio update rate in a nonlinear manner, adapting to the progressive stages of training. To achieve this, we employ a Sigmoid function to dynamically compute the step size $\eta$, thereby controlling the update speed of the masking ratio throughout the training process. The specific formula is given as follows:

$$\eta = \eta_{min} + (\eta_{max} - \eta_{min}) \cdot (1 - \frac{1}{1+exp(-k \cdot (\frac{epoch}{total\_epochs} - midpoint))}) \tag{1}$$

Here, $\eta_{min} = 0.01$ and $\eta_{max} = 0.02$ represent the minimum and maximum values of the step size, respectively. The parameter $k$ controls the steepness of the Sigmoid function and is set to 10. The midpoint determines the center position of the step size variation during training and is set to 0.5 to ensure the symmetry of the function.

This nonlinear adjustment method offers two significant advantages. First, in the early stages of training, the step size varies more rapidly, allowing the model to quickly adapt to the learning of coarse-grained features. Second, in the later stages of training, the step size gradually decreases, enabling finer adjustments to the masking ratio and promoting the learning of fine-grained features.

(2) Loss-driven mask ratio update

This strategy enables the masking ratio to adaptively respond to the model's performance on the current task, facilitating dynamic feedback learning. At each training step, $\mu_t$ is dynamically updated based on the current loss $L_{current}$ and the moving average of recent losses $L_{mean}$. The specific formula for updating the masking ratio is as follows:

$$\mu_t = \begin{cases} \mu_t - \eta, & if\ L_{current} \geq L_{mean} \\ \mu_t + \eta, & if\ L_{current} < L_{mean} \end{cases} \tag{2}$$

When $L_{current} \geq L_{mean}$, it indicates that the model may be struggling during training, possibly due to detection failures or overfitting to noise. In this case, reducing the masking ratio helps the model refocus on learning fundamental features, thereby restoring convergence. Conversely, when $L_{current} < L_{mean}$, it suggests that the model has made progress in learning specific features. Increasing the masking ratio appropriately raises the difficulty level, further enhancing the model's ability to adapt to more challenging features.

Overall, through the synergy of these two mechanisms, our feature masking strategy dynamically adjusts the learning focus at different training stages. This optimizes the model's understanding of feature representation in terms of both depth and breadth. Conceptually, it better balances the trade-off between progressive feature difficulty and model generalization capability.

### *3.2.2 Reconstruction*

After masking the features, the encoder processes the masked feature maps. In the output features of the encoder, we share masked queries to supplement the masked regions and pass the processed features to the MAE decoder for reconstruction. Considering that the last-layer feature map $F_L$ contains the most complete semantic information, we perform reconstruction only on the last layer. This reduces computational complexity while accelerating model convergence. The final layer of the decoder employs a linear projection, with the number of output channels matching that of $F_L$.

Finally, to supervise the reconstruction loss $L_{mask}$, we compute the mean squared error (MSE) between the reconstructed feature $\hat{F}_L$ and the original feature $F_L$. Additionally, the overall loss of the final student model is the sum of $L_{mask}$ and the teacher model loss $L_{teach}$. The loss functions are defined as follows:

$$L_{mask} = L_{MSE}(\hat{F}_L, F_L) \tag{3}$$

$$\mathcal{L} = L_{mask} + L_{teach} \tag{4}$$

## 3.3 Variance Feedback Smoothing Threshold Strategy

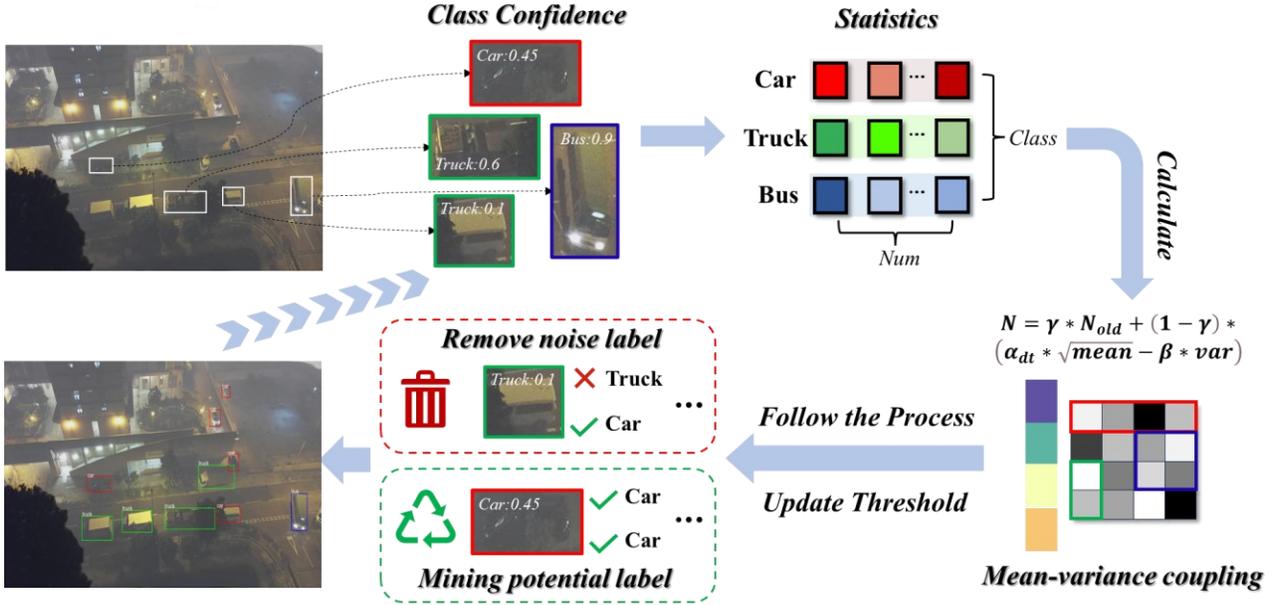

**Fig. 5.** The overall process of VFST.

The quality of pseudo-labels directly affects the effectiveness of the teacher model in guiding the student model and is closely related to the confidence threshold. However, due to the potential presence of noisy data or distribution bias in pseudo-labels, using a fixed threshold for each class or applying simple filtering may limit both the utilization and reliability of pseudo-labels during training. To address this issue, this paper proposes the variance feedback smoothing threshold strategy to enhance the quality and utilization of pseudo-labels for each class, as illustrated in Figure 5. The core formula of this method is as follows:

$$N = max\left(min\left(\gamma * N_{old} + (1-\gamma) * \left(\alpha_{dt} * \sqrt{mean} - \beta * var\right), max_{dt}\right), min_{dt}\right) \quad (5)$$

The threshold $N$ is determined by the historical threshold $N_{old}$ and the current statistical information, including the mean confidence score $\sqrt{mean}$ and the confidence variance $var$ for each category. The parameter $\gamma$ is a dynamic smoothing coefficient that balances historical and current information, ensuring both stability and gradual adjustment. The constraints $max_{dt}$ and $min_{dt}$ prevent the threshold from exceeding a reasonable range. The hyperparameters are set as $\alpha_{dt} = 0.5$ and $\beta = 0.2$, which can be adjusted as needed. The following sections provide a detailed explanation of the variance feedback mechanism and the dynamic smoothing strategy in VFST.

First, regarding the variance feedback mechanism, the threshold update strategy proposed by Zhao et al. (2023) utilizes the mean confidence score to reflect the overall trend of each class's pseudo-label confidence. However, this approach fails to capture fluctuations in pseudo-label distributions and is susceptible to outliers and noisy pseudo-labels, making the selection process less robust. Variance can serve as a measure of uncertainty. In threshold updates, incorporating a variance penalty term allows for stricter filtering of low-confidence pseudo-labels, preventing selection failures caused by high-variance scenarios while also identifying potentially useful pseudo-labels. The mean confidence score acts as a global confidence indicator, and when combined with variance, it provides a more comprehensive characterization of pseudo-label distribution properties. This dynamic balance between pseudo-label quality and quantity enhances the reliability of pseudo-label selection and improves overall training performance.

Second, regarding the dynamic smoothing mechanism ($\gamma$), in the early stages of training, the model's predictions may include unreliable pseudo-labels, leading to significant fluctuations in statistical results. By applying smoothing control, excessive threshold variations caused by noise can be mitigated. In the later stages, when the pseudo-label distribution stabilizes, increasing the weight of current information allows the model to better leverage high-confidence pseudo-labels for optimization. The specific implementation formula is as follows.

$$\gamma = \frac{1}{1 + exp\left(-\alpha_{dt} * \frac{current\_iter}{total\_iters}\right)} \tag{6}$$

In the early stages of iteration, since the pseudo-label distribution has not yet stabilized and the model's prediction quality remains low, relying on the historical threshold $N_{old}$ helps prevent the introduction of unreliable pseudo-labels. During the mid-iteration phase, as training progresses, the smoothing coefficient $\gamma$ gradually transitions from 1 to 0, shifting the threshold dependency from historical values to current statistical information. This enhances the model's adaptability to the evolving pseudo-label distribution. In the later stages, as the current iteration approaches the total number of iterations, the threshold increasingly aligns with the current statistical results, better reflecting the true distribution of pseudo-labels. Overall, this strategy enables continuous threshold updates throughout the training process, dynamically adapting to changes in the pseudo-label distribution. As a result, pseudo-label selection becomes more precise and reliable, ultimately improving object detection accuracy.

It is worth noting that from equation 5, we can observe that when the variance $var$ increases, the threshold decreases compared to the case without variance penalty. This implies that the model relaxes the selection criteria for pseudo-labels. This observation raises an important question: When the model exhibits high uncertainty in predicting a certain category (indicated by a larger variance), why not further increase the threshold to retain only high-confidence pseudo-labels? Instead, why is the threshold relaxed? Would this not introduce more uncertain pseudo-labels, potentially compromising the stability of model training?

It is important to explain why, when the model exhibits high uncertainty in predicting a certain category, VFST applies a variance penalty constraint on the threshold rather than directly increasing it. This design is based on the following considerations:

(1) Variance reflects distribution complexity: A large variance indicates that the confidence distribution of pseudo-labels for that category contains both high- and low-confidence samples (e.g., [0.1, 0.8]). If the threshold were directly increased (e.g., to 0.8), all pseudo-labels might be filtered out, preventing the model from learning features for that category.

(2) Edge filtering in the distribution: VFST lowers the threshold while incorporating a variance penalty to gradually refine pseudo-label selection. For instance, given a confidence sequence [0.2, 0.5, 0.6, 0.9], when variance is high, the threshold may decrease to 0.5. Low-confidence labels (e.g., 0.2) are filtered out due to their position at the edge of the distribution, while mid-confidence labels (e.g., 0.5) may correspond to real samples (such as boundary or hard samples). Allowing these samples to participate in training enhances the model's ability to capture intra-class diversity.

(3) Dual safeguard for training stability: The constraints $max_{dt}$ and $min_{dt}$ limit the range of threshold adjustments, preventing instability due to sudden variance fluctuations. In the later stages of training, as model predictions become more stable and the pseudo-label distribution converges, the influence of the variance penalty diminishes. Consequently, the threshold naturally converges to a reasonable range (e.g., approaching the mean), gradually reducing the inclusion of uncertain labels.

## 4 Dataset and Experiment Details

### *4.1 Dataset*

In the fundamental validation experiments, we used two datasets: one is the synthetic dataset from the HazyDet Dataset, and the other is the Real-Hazy Drone Detection Testing Set (RDDTS). These datasets were introduced by Feng et al. (2024a) and represent the first benchmark datasets designed for object detection under adverse conditions in remote sensing scenarios. The synthetic dataset consists of clear images paired with their corresponding synthesized foggy images. It is divided into a training set (8,000 image pairs with 264,511 objects), a validation set (1,000 image pairs with 34,560 objects), and a test set (2,000 image pairs with 65,322 objects). The RDDTS dataset is specifically used to evaluate the generalization ability of models in real-world scenarios. It contains only 600 real foggy images with corresponding annotations, including a total of 19,296 objects. These objects exhibit multi-scale variations, color and texture differences, as well as structural and appearance diversity within and across categories. Representative sample images are shown in Figure 6. Furthermore, we provide per-category instance statistics from the dataset to more fully support our observation that the model achieves higher gains on under-represented categories, as shown in Figure 7.

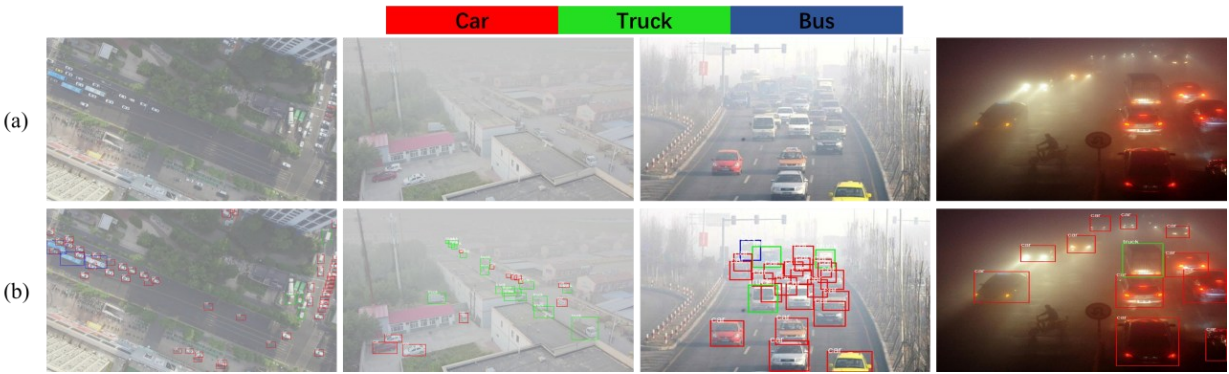

**Fig. 6** Some samples from the HazyDet dataset. (a) Original UAV image; (b) Ground truth for object detection.

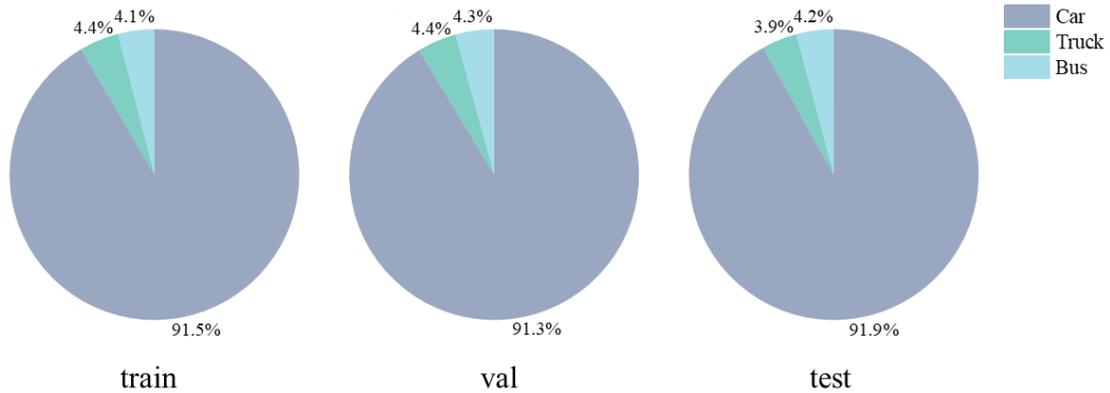

**Fig. 7** Statistics of instances for different categories in the HazyDet dataset.

## *4.2 Experiment Details*

All experiments in this study are conducted on a Linux server equipped with four NVIDIA GeForce RTX 4090 24GB GPUs. We use the Adam optimizer with an initial learning rate set to 2e-4 and a batch size of 8. The MAE employed a two-layer asymmetric decoder. Data augmentation techniques included random flipping, Gaussian blur, and color jittering. Additionally, to help the student model escape local optima, we adopt the Selective Retraining Strategy (SRS) following the approach of Zhao et al. (2023). In SF-TMAT, the student model's backbone and encoder parameters are initialized using the source training weights after a specified number of training epochs.

## 5 Experiment Results and Discussion

## *5.1 Basic Experiments on the HazyDet Dataset*

## *5.1.1. Quantitative Analysis*

To verify the superiority of the proposed method on benchmark dataset, we conducted comprehensive comparisons with state-of-the-art UDA methods, including SFA (Wang et al., 2021), O2net (Gong et al., 2022), MRT (Zhao et al., 2023), and RST (Han et al., 2024), as well as supervised methods such as Sparse RCNN (Sun et al., 2023), Dynamic RCNN (Zhang et al., 2020), Faster RCNN (Ren et al., 2017), FCOS

(Tian et al., 2019), YOLOX (Ge et al., 2021) and DAB DETR (Liu et al., 2022a). Table 1 presents the performance of these methods on the test set, while Figure 6 illustrates the parameter count and complexity of different UDA methods.

(1) *Overall performance comparison:* The proposed method (Ours) achieves an mAP$_{50}$ of 66.9%, significantly outperforming the best existing UDA methods, SFA (65.8%) and RST (64.6%). Compared to supervised learning methods (Sup.), our approach surpasses the best-performing Dynamic RCNN (65.0%) by 1.9 percentage points and achieves a remarkable performance improvement of +32.3% over the source-only model Def DETR (34.6%). These results demonstrate that the proposed method effectively mitigates cross-domain feature distribution shifts and even outperforms supervised methods.

**Table 1.** Quantitative results evaluated with respect to mAP$_{50}$ on HazyDet Test-set. The best quality score is **highlighted**.

| Methods | Type | AP on HazyDet Test-set | | | |
|---|---|---|---|---|---|
| | | Car | Truck | Bus | mAP$_{50}$ |
| Sparse RCNN | *Sup.* | 51.1 | 22.6 | 49.9 | 41.2 |
| Dynamic RCNN | *Sup.* | 79.3 | 39.7 | 75.9 | 65.0 |
| Faster RCNN | *Sup.* | 56.3 | 30.5 | 59.3 | 48.7 |
| YOLOX | *Sup.* | 53.1 | 23.0 | 51.2 | 42.3 |
| FCOS | *Sup.* | 54.4 | 27.1 | 56.2 | 45.9 |
| DAB DETR | *Sup.* | 70.0 | 28.4 | 69.7 | 56.0 |
| Def DETR | *Source* | 48.8 | 13.6 | 41.4 | 34.6 |
| SFA | *UDA* | **82.5** | 42.2 | 72.8 | 65.8 |
| O2net | *UDA* | 75.6 | 43.3 | 73.2 | 64.0 |
| MRT | *UDA* | 77.8 | 37.5 | 77.3 | 64.2 |
| RST | *UDA* | 82.1 | 41.3 | 70.4 | 64.6 |
| Ours | *UDA* | 77.9 | **44.9** | **78.1** | **66.9** |

(2) *Fine-grained category performance analysis:* SF-TMAT achieves the best results on the minority classes Bus (78.1% AP) and Truck (44.9% AP), significantly outperforming the next-best method. The primary reason for this phenomenon is that our proposed VFST thresholding strategy can more robustly handle the sparser and potentially noisier pseudo-labels found in minority classes; it effectively suppresses predictions with high uncertainty while simultaneously uncovering and preserving valuable potential true positive labels that might be filtered out by simpler thresholding strategies, especially for classes with inherently limited data samples. In contrast, on the majority class, Car, although SF-TMAT (77.9% AP) does not achieve the absolute best performance (compared to, e.g., SFA's 82.5% AP), it still demonstrates strong competitiveness. This performance profile is primarily attributed to the optimization focus of our method on targeting adverse conditions and class imbalance. Specifically, VFST's stricter noise suppression mechanism, while benefiting overall precision, might filter out some medium-to-low confidence pseudo-labels for the high-density Car category, potentially impacting its recall slightly. Concurrently, the DSFMA strategy focuses on learning features with general robustness to adverse environments (such as haze); although this benefits all classes, its relative advantage becomes more pronounced for categories like Bus and Truck which face the dual challenges of data sparsity and adverse environmental conditions. Therefore, this observed performance distribution is a natural consequence of SF-TMAT's successful optimization for the best overall mAP (66.9%) and its enhanced comprehensive detection capability under challenging conditions involving class imbalance and adverse environments.

(3) *Accuracy-complexity trade-off evaluation:* As shown in Figure x, the proposed method has a parameter size of 39.82M, which is identical to MRT but superior to all other methods. In terms of computational complexity (103.28G), our method is on par with MRT and O2net, while outperforming the remaining approaches. It is important to explain why our method shares the same parameter count and computational cost as MRT. This is because both methods are based on the Deformable DETR detector and do not modify its original architecture; instead, our contributions focus on threshold strategy improvements. To better evaluate the balance between accuracy and computational complexity, the subfigure in Figure 8 (with a blue background) presents the accuracy-to-complexity ratio (mAP/GFLOPs) for different UDA methods. Our method achieves the best accuracy-to-complexity ratio, indicating that it offers a superior balance between detection performance and computational cost.

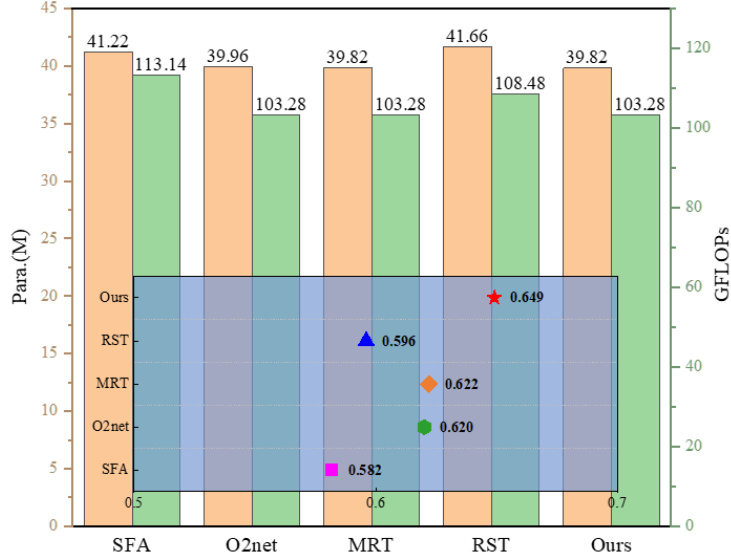

**Fig. 8.** Bar chart of the number of parameters and GFLOPs for each method, and a dot plot of the accuracy-complexity ratio.

Overall, the proposed method not only achieves state-of-the-art accuracy in UAV cross-domain detection tasks but also maintains an excellent balance between computational cost and performance. Notably, it demonstrates superior performance on low-sample categories, such as Truck and Bus, further highlighting its effectiveness in handling underrepresented targets.

## *5.1.2. Qualitative Analysis*

To better illustrate the superiority of the proposed method in UAV object detection under adverse conditions, this section analyzes common challenging scenarios in object detection. We compare detection results from SF-TMAT, the MRT method (which also incorporates a threshold adjustment mechanism), and RST, a method specifically designed for object detection in remote sensing imagery (as shown in Figures 10 and 11). The evaluation focuses on five challenging scenarios: small objects, occlusion, complex backgrounds, dense objects, and nighttime scenes. Furthermore, to demonstrate the cross-domain feature alignment capability of the proposed method, we employ t-SNE for feature dimensionality reduction and visualization (Figure 8). Additionally, we perform kernel density estimation (KDE) to provide a more intuitive representation of cross-domain alignment results (Figure 8).

(1) *Cross-domain distribution alignment analysis:* Based on the visualizations in Figures 9 and 10, the effectiveness of cross-domain distribution alignment is significantly improved: First, in Figure 9, the detector trained solely on the source domain (Figure 9(a)) exhibits a scattered feature distribution, with a clear domain gap. In contrast, our proposed SF-TMAT (Figure 9(b)) effectively bridges this gap, resulting in a more compact feature distribution. Second, in Figure 10(b), the contour alignment between the source and target domains after domain adaptation is notably improved compared to the detector trained only on the source domain (Figure 10 (a)). Before adaptation, the source domain contours in Figure 10 (a) form densely closed rings in the central region, indicating a high-density kernel, while the target domain contours appear dispersed and shifted, creating an asymmetric bimodal structure. After adaptation, the primary contours from both domains spatially overlap, forming a continuous golden overlap zone (red-blue fusion region), indicating that the geometric centers of the feature distributions have become aligned. Notably, the previously isolated secondary peak in the target domain (lower right region in Figure 10 (a)) is suppressed after adaptation. Its contours shrink toward and merge with the primary peak of the source domain, further demonstrating that cross-domain semantic ambiguity has been effectively resolved.

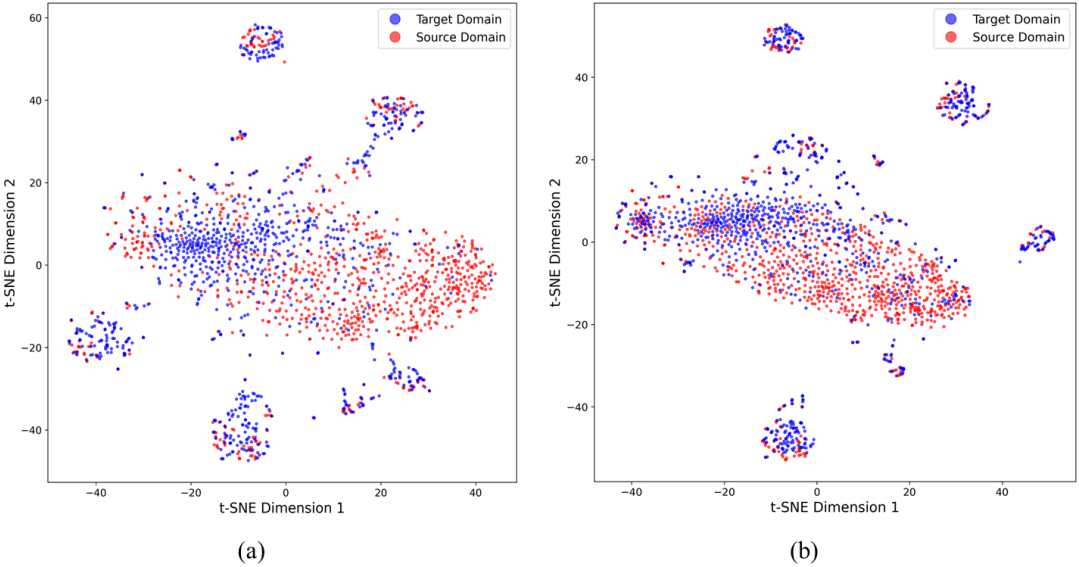

**Fig. 9.** Feature distribution visualization by t-SNE (Van der Maaten and Hinton, 2008) before and after domain adaptation. (a) original distribution before adaptation; (b) aligned distribution after the proposed adaptation.

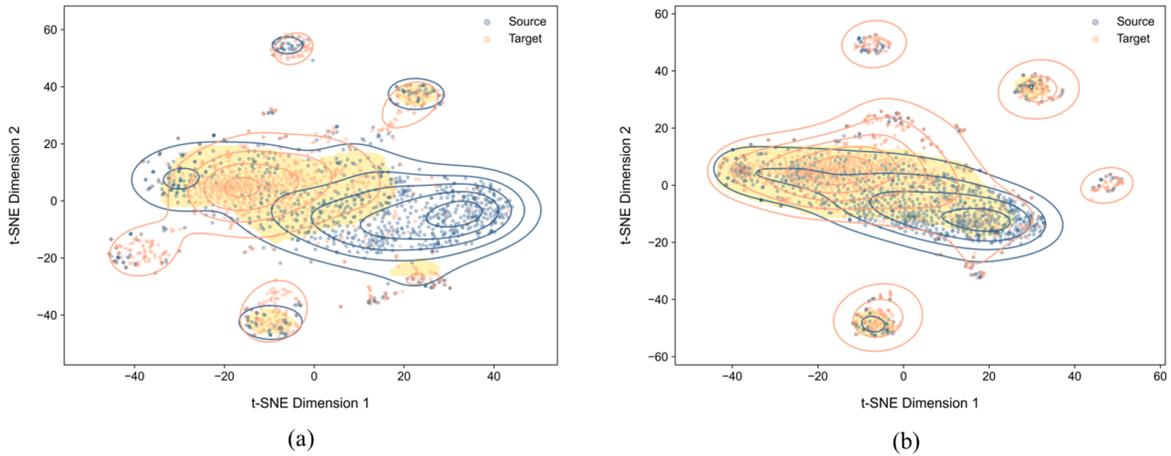

**Fig. 10.** KDE on t-SNE embeddings. (a) KDE on source data features; (b) KDE on aligned features.

(2) *Challenging daytime scene object detection:* In the "small object" scenario, despite the small size of targets and the presence of fog interference, our method effectively captures distant small objects in the image. While MRT is also able to detect distant cars, our method provides detections that are closer to ground truth, whereas MRT produces a large number of false detections for buses. In contrast, RST is less sensitive to distant small objects, leading to a significant number of missed detections. In the "occlusion" scenario, as seen in the images, various objects partially occlude vehicles, causing feature discontinuities that make learning more challenging. However, our method is able to accurately distinguish occluded targets and predict their locations with high precision. In the "complex background" scenario, the diverse structures of toll stations, the dense arrangement of different object categories, and fog interference make even human recognition difficult. Under these conditions, our method is able to detect the maximum number of objects across different categories, whereas other methods suffer from a significantly higher number of missed detections. In the "dense object" scenario, where multiple objects are tightly clustered and distant targets are heavily obscured by dense fog, our method better mitigates the problem of missed detections due to object overlap, outperforming other methods in handling highly congested scenes.

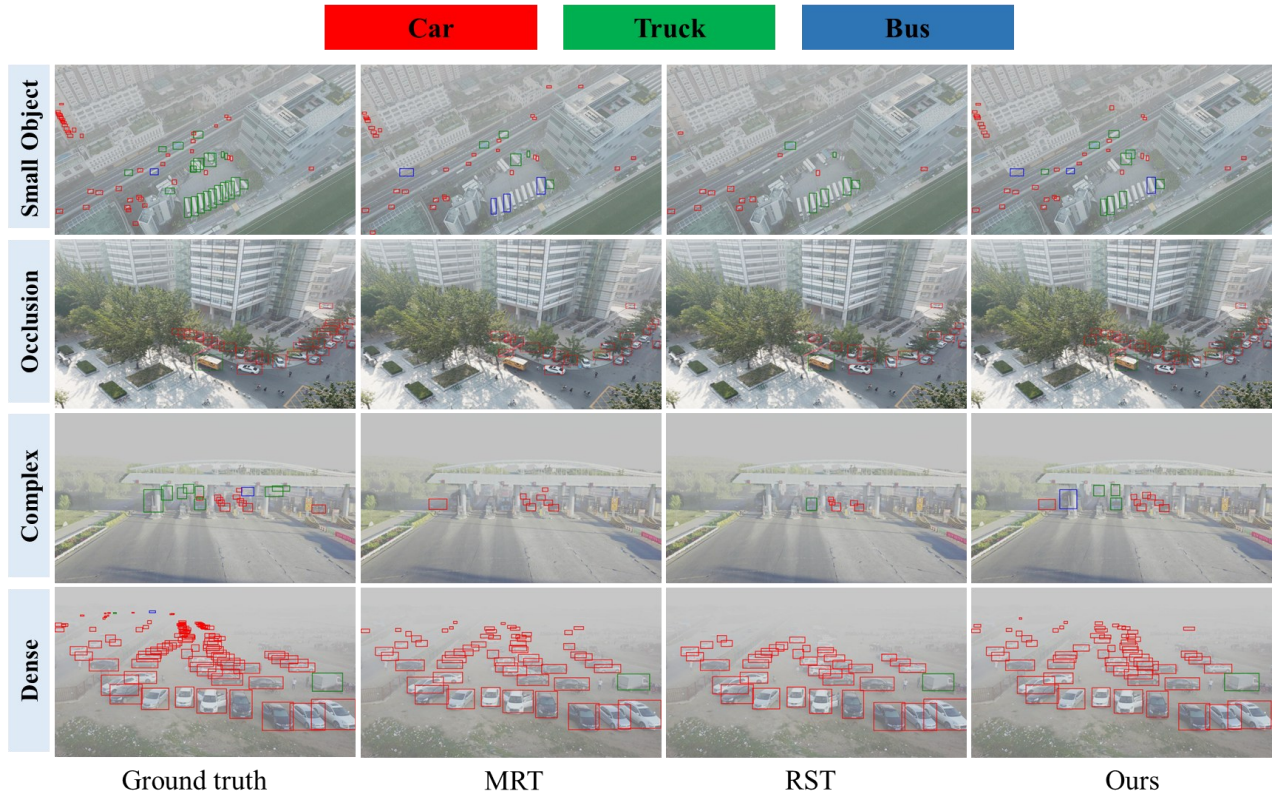

**Fig. 11.** Our detection results compared with the state-of-the-art methods. Different categories are marked with different colors. The confidence threshold for visualization is set to 0.5. Each row represents the detection results for different challenging daytime scenarios. Other descriptions represent their respective methods.

(3) *Object detection in nighttime scenes:* Figure 12 presents the detection performance of different methods under low-light and adverse environmental conditions. First, in challenging nighttime scenarios, vehicle headlights (as seen in the first row of detection images) often interfere with the features learned by models trained on daytime scenes, potentially leading to localization drift or missed detections. Compared to other methods, the proposed approach achieves more accurate localization of cars and provides detections closer to the ground truth for densely packed distant vehicles. Second, under low-light conditions, partial occlusion further blurs object features, making detection even more difficult. This effect can be observed in the detection results in the middle and last rows of Figure 12. Compared to other methods, our approach achieves higher completeness in detected objects, while other methods suffer from a large number of missing occluded and distant targets, particularly RST, which exhibits the most severe performance degradation.

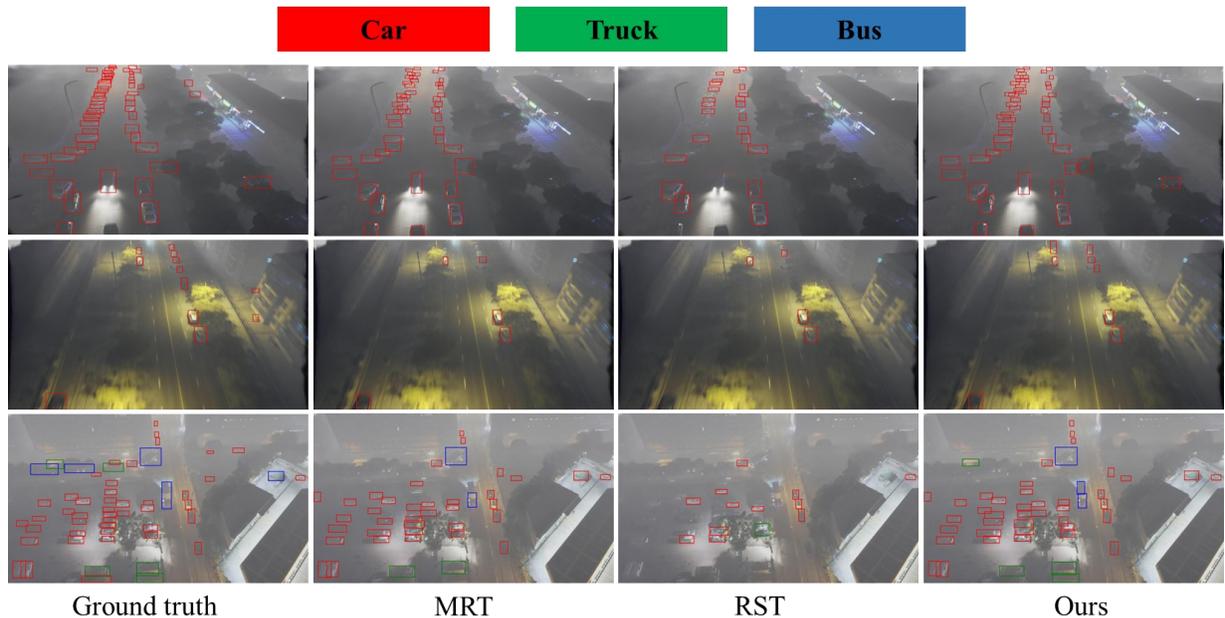

**Fig. 12.** Our detection results compared with the state-of-the-art methods in challenging nighttime scenarios. Different categories are marked with different colors. The confidence threshold for visualization is set to 0.5. Other descriptions represent their respective methods.

Overall, through qualitative analysis across various challenging scenarios, our method demonstrates significant advantages in diverse adverse environments. Whether in small object detection, occlusion, complex backgrounds, or dense object scenes, our approach maintains a high level of detection completeness. In nighttime environments, our method is able to preserve high detection accuracy even under low-light conditions and strong interference, further showcasing its robustness in extreme scenarios.

## 5.2 Performance in Real Scenes

Validating the robustness and generalization capability of an algorithm in real-world scenarios is a key step in assessing its practical engineering value. Compared to controlled experimental environments, real-world conditions are often more complex and adverse, imposing higher demands on the generalization ability of object detectors. Therefore, in this section, we conduct quantitative evaluations and qualitative visualizations based on the real-world dataset RDDTS. It is important to note that the experimental results in this section are obtained by directly applying the trained weights from the HazyDet dataset to the new dataset, without additional fine-tuning.

First, as shown in Table 2, the overall accuracy performance of different methods is presented. Our proposed method achieves the best performance even when directly transferred to real-world scenarios, demonstrating its superior transferability compared to other state-of-the-art approaches. Second, to provide a more intuitive understanding of the new challenges posed by real-world conditions, we select three representative adverse weather scenarios from the dataset for visualization, as shown in Figure 13. These include hazy weather, dense fog, and snowy conditions.

**Table 2.** Quantitative results evaluated with respect to $mAP_{50}$ on RDDTS. The best quality score is **highlighted**.

| Methods | SFA | O2net | MRT | RST | Ours |
|---|---|---|---|---|---|
| $mAP_{50}$ | 36.6 | 36.9 | 35.5 | 37.3 | **37.7** |

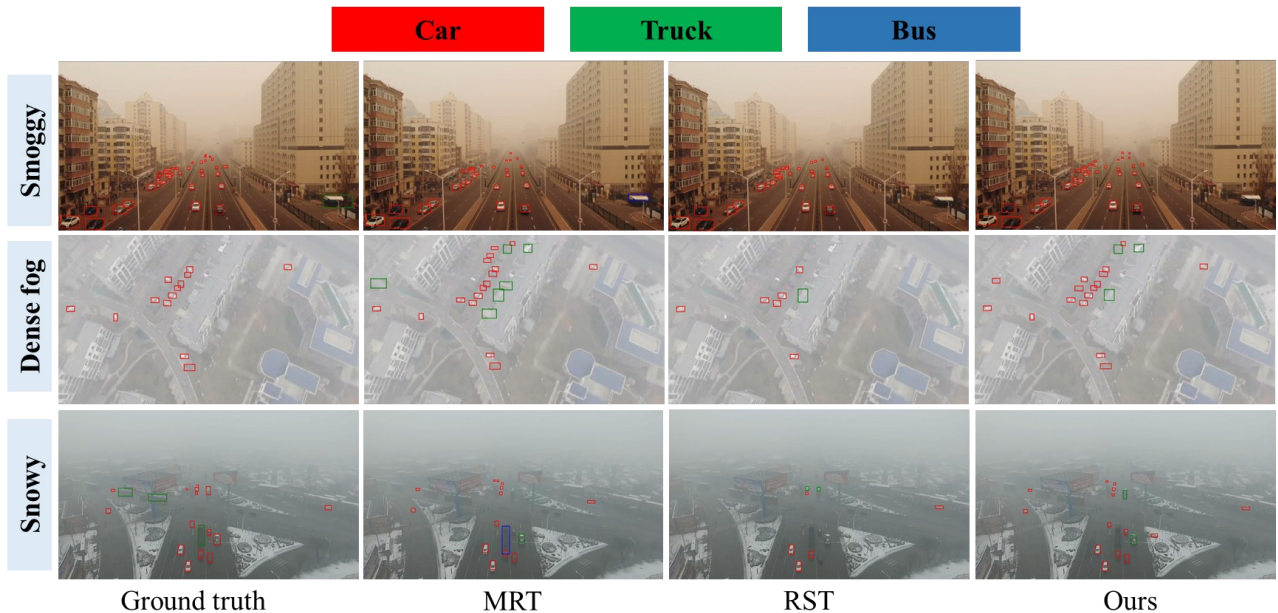

**Fig. 13.** Our detection results compared with the state-of-the-art methods. Different categories are marked with different colors. The confidence threshold for visualization is set to 0.5. Each row represents the detection results for different adverse scenarios. Other descriptions represent their respective methods.

From the first row of detection results, it can be observed that hazy weather alters the original color tones of the scene, and the presence of fog further increases the difficulty of accurate localization. However, compared to other methods, our approach detects more objects, especially distant small targets, demonstrating superior robustness in such conditions. The second row illustrates the impact of dense fog on object detection. The combination of dense fog and UAV's high-altitude perspective causes object features to become blurred, increasing the likelihood of confusion between background structures and target features, leading to false detections. As shown in the figure, all methods exhibit some degree of false positives on background structures, with MRT being the most affected. While our method has a higher false detection rate than RST, it achieves a much higher accuracy in correctly detecting targets, closely matching the ground truth. The third row presents detection results in a snow-and-fog scenario. It is evident that all methods fail to detect buses, as snow accumulation on vehicles creates a strong feature discrepancy between the observed and learned representations. Additionally, while MRT and our method produce competitive results, our approach outperforms in terms of accuracy and exhibits better localization contours.

## 5.3 Spectral Variability Effect Analysis

In remote sensing imagery, spectral variation is a widespread and unavoidable phenomenon, typically caused by factors such as illumination changes, terrain features, atmospheric interference, and material properties (Hong et al., 2019). This variation not only affects image quality, making analysis more challenging, but also poses obstacles to subsequent processing tasks, including image segmentation, classification, and object detection (Chen et al., 2023). In the context of UAV-based object detection, spectral variations can directly impact the discriminative features of targets, potentially leading to degraded detection performance. To thoroughly investigate the influence of spectral variation on UAV object detection, this section explores different simulated conditions, including illumination changes, Gaussian noise, and spectral artifacts. Representative examples of the simulated data are shown in Figure 14.

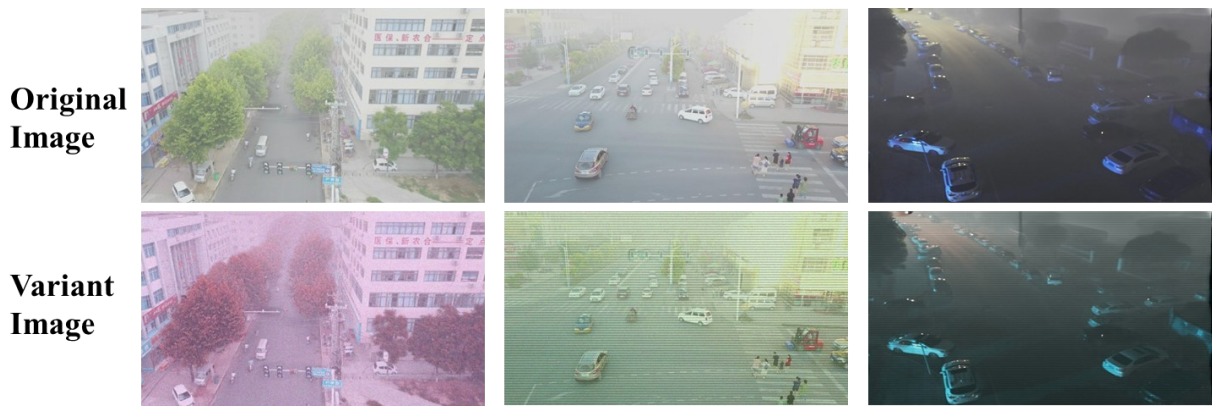

**Fig. 14.** Example samples after simulating spectral variations. The first row shows the original image, and the second row shows the simulated image after variation.

Similarly, we directly applied the pre-trained weights of each method to the HazyDet dataset with simulated spectral variations. Figure 15 records the accuracy changes for different methods. It can be observed that all methods experience performance degradation when applied to the simulated data, with SFA, O2net, and RST showing the most significant declines. Although our method exhibits slightly higher performance degradation compared to MRT, it still achieves the highest AP for each category as well as the highest mAP. Additionally, we utilized box plots to visualize the mAP distribution across categories for MRT and our method. As shown in Figure 16, both methods demonstrate competitive performance, but our approach maintains a slight advantage, outperforming MRT in mAP distribution stability and median accuracy.

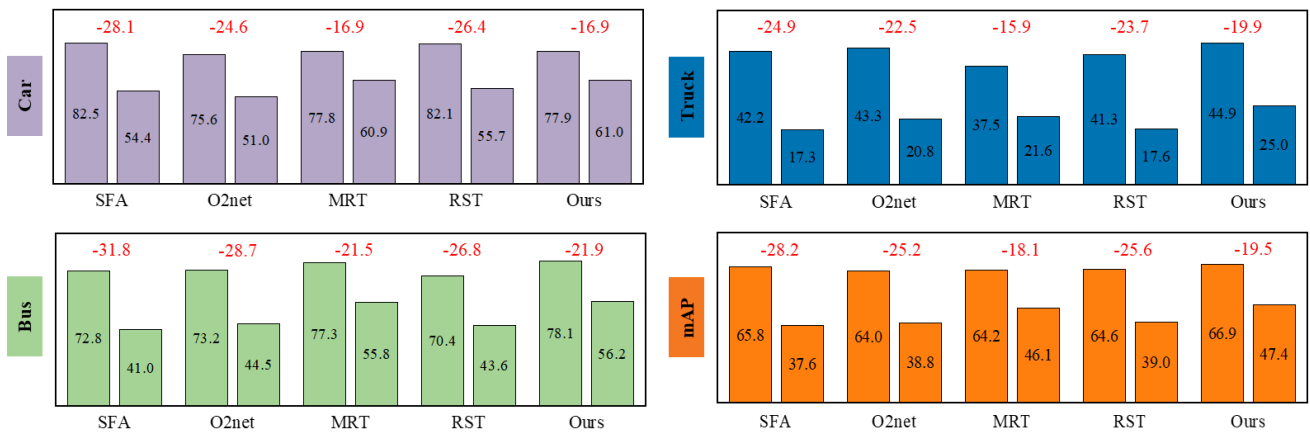

**Fig. 15.** The performance metrics of each method on data before and after spectral variation. This includes four bar charts, where the left bars represent the metrics obtained by each method before spectral variation (consistent with the results in Table 1), and the right bars represent the metrics obtained after spectral variation. The red text at the top of the bars indicates the magnitude of change in the metrics before and after variation.

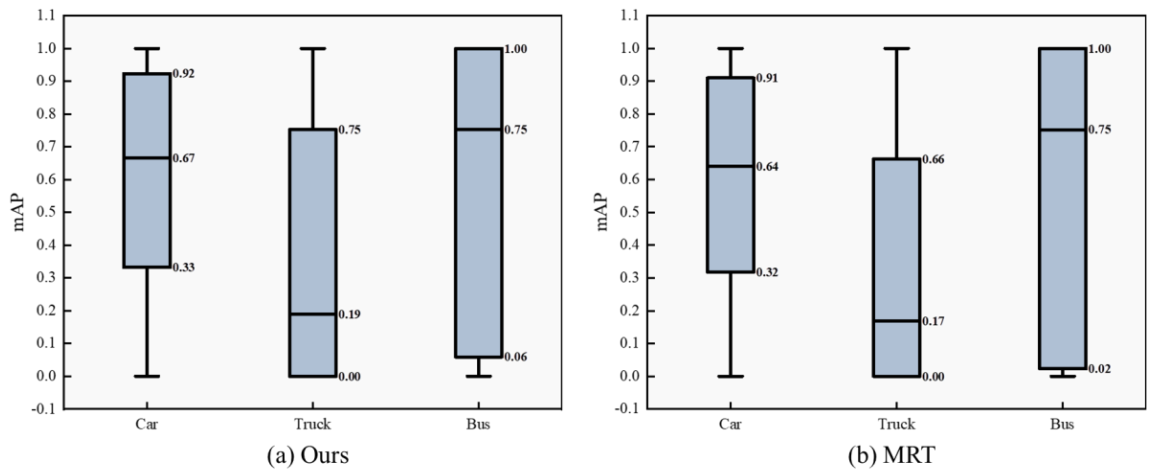

**Fig. 16.** Box plot of the metric distribution on post-variation data for our method and MRT.

## 5.4 Robustness Verification and Analysis

This section presents an experimental evaluation to validate the robustness of our method. The experiments were conducted using the DroneVehicle dataset proposed by Sun et al. (2022), which contains RGB and infrared images captured in urban roads, parking lots, and residential areas, with a predominance of nighttime imagery. This dataset is suitable for testing algorithmic performance under diverse environmental conditions. Since our study focuses on RGB imagery, we exclusively utilized the RGB images from the dataset and restricted our analysis to its validation subset, comprising 1,469 RGB images at a resolution of $640 \times 512$. Consistent with standard evaluation protocols, all comparative methods were tested using their pre-trained weights without additional retraining.

Table 3 and Figure 17 document the performance metrics and visualization results of all evaluated methods, respectively. Both quantitative and qualitative outcomes demonstrate that our method maintains superior performance on the new dataset. The visual detection results reveal that our approach exhibits fewer false positives and false negatives across all scenarios compared to other methods, with precise localization of targets. This robust performance provides empirical evidence for the substantial environmental adaptability and operational robustness of the proposed framework.

**Table 3.** Quantitative results evaluated with respect to $mAP_{50}$ on DroneVehicle dataset. The best quality score is **highlighted**.

| Methods | SFA | O2net | MRT | RST | Ours |
|---------|-----|-------|-----|-----|------|
| $mAP_{50}$ | 39.1 | 44.1 | 43.8 | 42.9 | **44.8** |

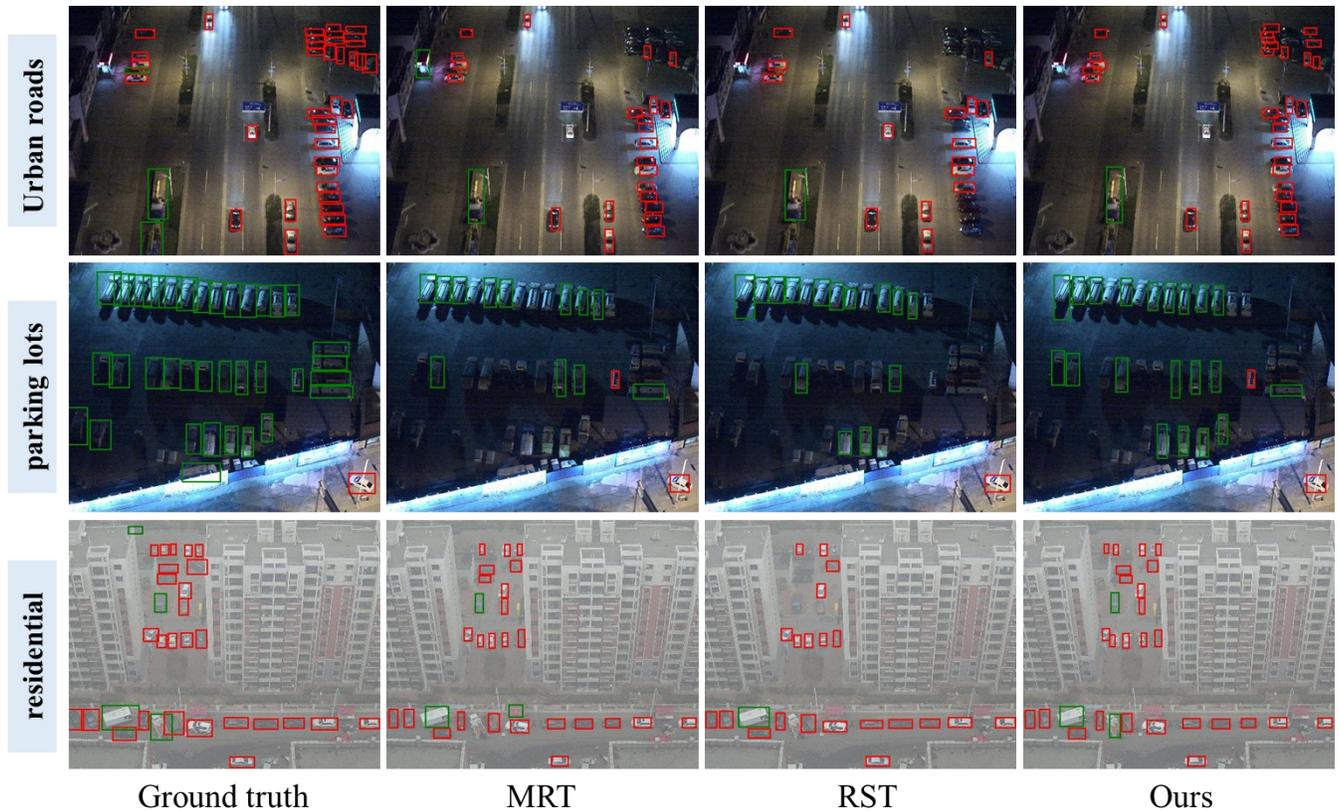

**Fig. 17.** Our detection results compared with the state-of-the-art methods. Different categories are marked with different colors. The confidence threshold for visualization is set to 0.5. Each row represents the detection results for different scenarios. Other descriptions represent their respective methods.

## *5.5 Ablation Study*

## *5.5.1. Basic Verification*

To comprehensively evaluate the effectiveness of the proposed module, we conducted detailed ablation experiments on the HazyDet dataset. Specifically, the following configurations are implemented:

• $A_0$: Trained exclusively on clear images and directly applied to object detection in adverse scenarios.

• $A_1$: Removed the teacher framework and adopted a fixed threshold, testing solely with weights obtained from cross-domain training.

• $A_2$: Fixed the mask ratio of the autoencoder during cross-domain training.

• $A_3$: Fixed the pseudo-label filtering threshold in the teaching process.

As shown in Table 4, training solely on source domain data fails to adequately address the challenges of the target domain, resulting in a significant performance decline compared to the full configuration. By incorporating cross-domain training and the proposed DSFMA technique, $A_1$ achieves a performance leap to 59.5, marking an absolute improvement of 24.9 points. This demonstrates that cross-domain training effectively enhances the model's adaptability. Furthermore, by comparing the full configuration $A_4$ with $A_2$ and $A_3$, it is evident that the removal of either DSFMA or VFST leads to a performance degradation. This confirms the substantial contribution of the two core techniques proposed in this study to improving the final object detection performance.

**Table 4.** Ablation experiments conducted on the HazyDet dataset. The best score is **highlighted**.

|       | Source | AT | DSFMA | VFST | mAP$_{50}$ |
|-------|--------|----|-------|------|------------|
| $A_0$ | ✓      |    |       |      | 34.6       |
| $A_1$ | ✓      |    | ✓     |      | 59.5       |
| $A_2$ | ✓      | ✓  |       | ✓    | 63.5       |
| $A_3$ | ✓      | ✓  | ✓     |      | 65.4       |
| $A_4$ | ✓      | ✓  | ✓     | ✓    | **66.9**   |

Furthermore, Table 5 further illustrates the impact of removing key components on object detection performance under higher IoU thresholds. As shown in the table, the two proposed components demonstrate a more significant effect at higher IoU thresholds. In particular, for categories with a relatively small number of samples in the dataset, such as the Bus category, when the IoU is increased to 0.75, the accuracy improves by 9.3% and 5.8% compared to $A_2$ and $A_3$, respectively, where the components are removed. Additionally, at higher IoU thresholds, the performance gap between the full configuration method and the methods lacking components widens further. These results indicate that the proposed DSFMA and VFST components maintain a significant advantage even under more stringent conditions.

**Table 5.** Performance of the method under higher IoU thresholds after removing key components. The best score is **highlighted**.

| Num. | AP (IoU=0.75 / 0.50:0.95) | | | |
|---|---|---|---|---|
| | Car | Truck | Bus | mAP |
| $A_2$ | 33.1/38.1 | 22.8/22.1 | 45.8/43.2 | 33.9/34.5 |
| $A_3$ | 36.2/39.4 | 25.7/23.9 | 49.3/45.0 | 37.1/36.1 |
| $A_4$ | **39.1/41.0** | **26.1/25.1** | **55.1/48.2** | **40.1/38.1** |

Figure 18 illustrates the pseudo-label generation results after fixing the mask ratio and threshold. As shown in the figure, when the mask ratio is fixed, the model fails to effectively learn the reconstruction of features at varying granularities, leading to missed detections and inaccurate localization for certain categories in the pseudo-labels. Conversely, when the threshold is fixed, the model cannot dynamically adjust based on the quality and quantity of pseudo-labels during training, resulting in unreliable pseudo-labels or the exclusion of potentially valid ones. In contrast, the proposed method, empowered by DSFMA and VFST, generates pseudo-labels whose quantity and quality closely approximate the ground truth.

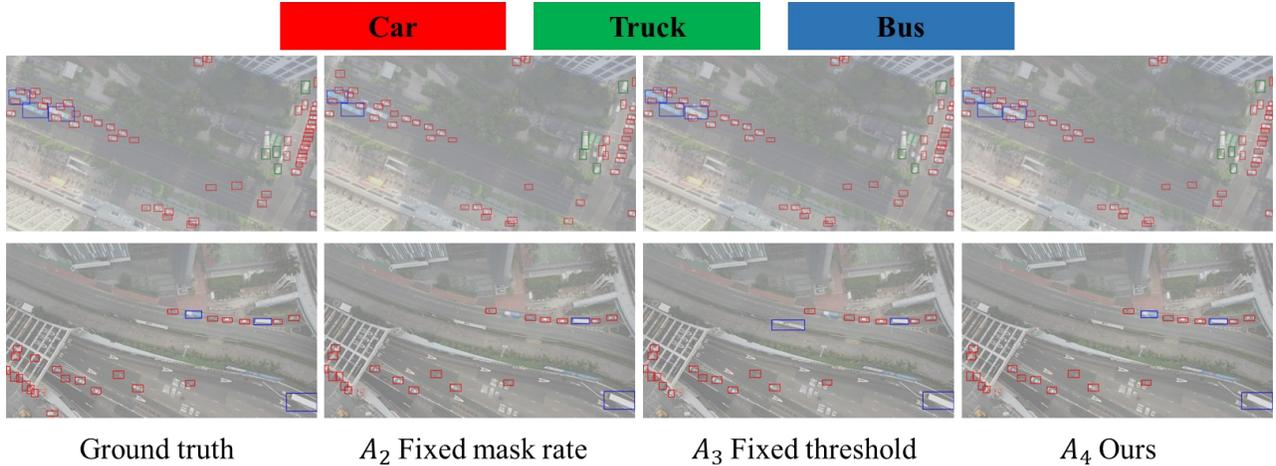

Ground truth     $A_2$ Fixed mask rate     $A_3$ Fixed threshold     $A_4$ Ours

**Fig. 18.** Detection results after removing the two key components of the proposed method. Different categories are marked with different colors. The confidence threshold for visualization is set to 0.5.

### 5.5.2. Feature Visualization Verification

To more intuitively observe the contribution of key components in the proposed method, this section visualizes the output features and presents them in the form of heatmaps. As shown in Figure 19, when

VFST is removed, the method exhibits a reduced level of fine-grained attention to potential target regions and a weakened feature response to some difficult-to-detect objects. This phenomenon may be attributed to the fact that without VFST, the method's ability to mine potentially useful labels is impaired, leading to reduced confidence in reliable pseudo-labels. Furthermore, after removing DSFMA, the feature response to target regions is noticeably weakened. This is because DSFMA is designed to encourage the model to learn different fine-grained features, and its removal results in a diminished ability of the model to effectively represent target features. Therefore, we conclude that VFST and DSFMA make practical and reasonable contributions to UAV object detection in adverse environments.

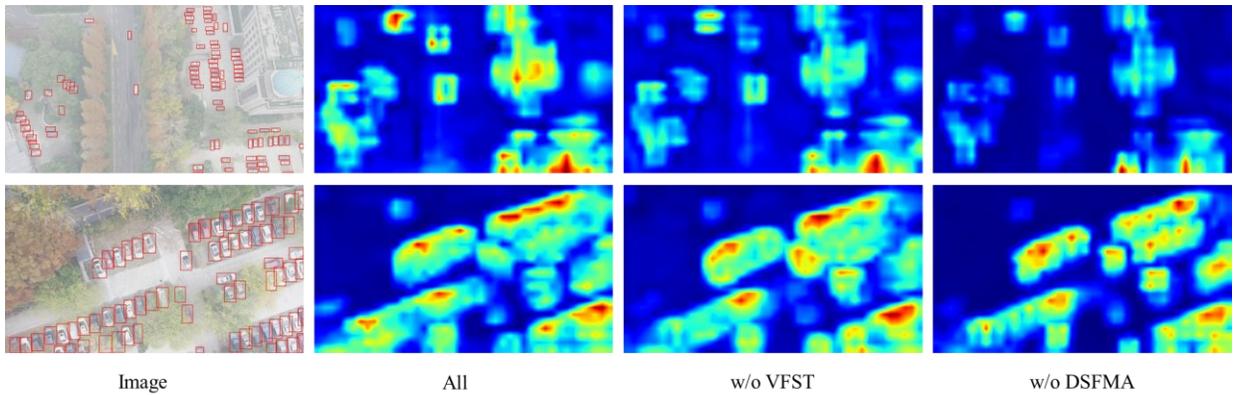

**Fig. 19.** Feature visualization heatmaps on the HazyDet dataset for the complete method and the method after removing key components.

## 6 Conclusions and Future Work

To address the gap in the application of UDA methods for UAV object detection under adverse scenes, we propose SF-TMAT, the first UDA method specifically designed for UAV object detection in such conditions. The core of this method consists of two components: DSFMA and VFST. DSFMA dynamically adjusts the step size based on the Sigmoid function and incorporates training progress and loss to enable dynamic feedback learning, adaptively updating the mask ratio. This strategy better balances the relationship between feature difficulty progression and the model's generalization ability. VFST, during the training phase, captures the fluctuation in pseudo-label distribution by statistically computing the average confidence and variance for each class, thereby dynamically adjusting the threshold to improve pseudo-label quality. To validate the effectiveness of the proposed method, we conducted extensive experiments, including baseline qualitative and quantitative evaluations, real-world scenario adaptability tests, comparisons before and after spectral variation, robustness experiments, and detailed ablation studies. In all experiments, SF-TMAT demonstrated a significant advantage and competitiveness compared to other state-of-the-art methods, providing a strong benchmark for UAV object detection in adverse scenes.

Although SF-TMAT has demonstrated superiority in most cases, the method suffers significant performance degradation when directly transferring the model to new data in some specific scenarios. As shown in Figure 20, it is evident that when faced with sufficiently low-altitude UAV perspectives (close to natural images), the detection performance of the method drastically declines. This is because the vast majority of data in UAV datasets are captured from medium to high-altitude perspectives, with very few or no images from ultra-low-altitude perspectives. As a result, the features learned by the method may be confined to the top plane or limited side-view information of the vehicles. Ultra-low-altitude perspectives make the targets more three-dimensional, containing more detailed features, which leads to the method failing to generalize well to these long-tail distribution data.

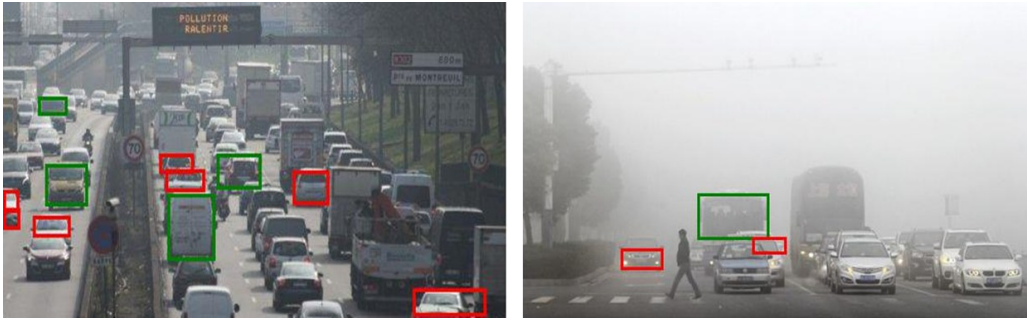

**Fig. 20.** The presentation of failure cases of the proposed method. The confidence threshold for visualization is set to 0.5.

In future work, we aim to further enhance the model's generalization ability, enabling it to better handle UAV imagery from different perspectives. Additionally, due to the limitations of the dataset categories, this study is restricted to the detection of specific vehicles. In future research, we hope to extend the method to be applicable in open scene scenarios.

## *Acknowledgments*

This work was supported in part by the National Natural Science Foundation of China (Grant No. 42301473), in part by the Key Research and Development Project of Sichuan Province (Grant No.

24ZDYF0633), in part by the Natural Science Foundation of Sichuan Province (Grant No. 24NSFSC2264), in part by the China Postdoctoral Science Foundation (Grant No. 2023M742884), in part by the Postdoctoral Innovation Talents Support Program (Grant No. BX20230299) The authors would like to thank the anonymous reviewers for their constructive and valuable suggestions on the earlier drafts of this manuscript.

## References

Appiah, E.O., Mensah, S., 2024. Object detection in adverse weather condition for autonomous vehicles. *Multimed. Tools Appl.* 83(9), 28235–28261.

Biswas, D., Tešić, J., 2024. Unsupervised domain adaptation with debiased contrastive learning and support-set guided pseudo labeling for remote sensing images. *IEEE J. Sel. Top. Appl. Earth Obs. Remote Sens.* 17, 3197-3210.

Cao, Y., Zhang, H., Lu, X., et al., 2024. Towards source-free domain adaptive semantic segmentation via importance-aware and prototype-contrast learning. *IEEE Trans. Intell. Veh.* Early Access.

Chen, H., Feng, D., Cao, S., et al., 2023. Slice-to-slice context transfer and uncertain region calibration network for shadow detection in remote sensing imagery. *ISPRS J. Photogramm. Remote Sens.* 203, 166-182.

Chen, M., Chen, W., Yang, S., et al., 2022. Learning domain adaptive object detection with probabilistic teacher. *arXiv:2206.06293*.

Chen, Y., Wang, W., Li, W., et al., 2018. Domain adaptive faster R-CNN for object detection in the wild. *Proc. IEEE/CVF Conf. Comput. Vis. Pattern Recognit. (CVPR)*, pp. 3339–3348.

Chen, Y., Ye, Z., Sun, H., et al., 2025a. Global-local fusion with semantic information-guidance for accurate small object detection in UAV aerial images. *IEEE Trans. Geosci. Remote Sens.* 63, 1–15.

Chen, Y., Yuan, X., Wang, J., et al., 2025b. YOLO-MS: rethinking multi-scale representation learning for real-time object detection. *IEEE Trans. Pattern Anal. Mach. Intell.* Early Access.

Du, B., Huang, Y., Chen, J., et al., 2023. Adaptive sparse convolutional networks with global context enhancement for faster object detection on drone images. *Proc. IEEE/CVF Conf. Comput. Vis. Pattern Recognit. (CVPR)*, pp. 13435–13444.

Feng, C., Chen, Z., Kou, R., et al., 2024a. HazyDet: open-source benchmark for drone-view object detection with depth-cues in hazy scenes. *arXiv:2409.19833*.

Feng, D., Chen, H., Liu, S., et al., 2023. Boundary-semantic collaborative guidance network with dual-stream feedback mechanism for salient object detection in optical remote sensing imagery. *IEEE Geosci. Remote Sens.* 61, 1-17.

Feng, J., Zhou, Z., Shang, R., et al., 2024b. Class-aligned and class-balancing generative domain adaptation for hyperspectral image classification. *IEEE Trans. Geosci. Remote Sens.* 62, 1–17.

Ge, Z., Liu, S., Wang, F., Li, Z., Sun, J., 2021. Yolox: Exceeding yolo series in 2021. arXiv preprint arXiv:2107.08430.

Gong, K., Li, S., Li, S., et al., 2022. Improving transferability for domain adaptive detection transformers. *Proc. 30th ACM Int. Conf. Multimedia*, pp. 1543–1551.

Guo, Y., Yu, H., Xie, S., et al., 2024. DSCA: a dual semantic correlation alignment method for domain adaptation object detection. *Pattern Recognit.* 150, 110329

Gupta, H., Kotlyar, O., Andreasson, H., et al., 2024. Robust object detection in challenging weather conditions. *Proc. IEEE/CVF Winter Conf. Appl. Comput. Vis. (WACV)*, pp. 7523–7532.

Han, J., Yang, W., Wang, Y., et al., 2024. Remote sensing teacher: cross-domain detection transformer with learnable frequency-enhanced feature alignment in remote sensing imagery. *IEEE Trans. Geosci. Remote Sens.* 62, 1–14.

Han, W., Chen, J., Wang, L., et al., 2021. Methods for small, weak object detection in optical high-resolution remote sensing images: a survey of advances and challenges. *IEEE Geosci. Remote Sens. Mag.* 9(4), 8–34.

Han, W., Li, J., Wang, S., et al., 2022. A context-scale-aware detector and a new benchmark for remote sensing small weak object detection in unmanned aerial vehicle images. *Int. J. Appl. Earth Obs. Geoinf.* 112, 102966.

He, K., Chen, X., Xie, S., et al., 2022a. Masked autoencoders are scalable vision learners. *Proc. IEEE/CVF Conf. Comput. Vis. Pattern Recognit. (CVPR)*, pp. 16000–16009.

He, M., Wang, Y., Wu, J., et al., 2022b. Cross domain object detection by target-perceived dual branch distillation. *Proc. IEEE/CVF Conf. Comput. Vis. Pattern Recognit. (CVPR)*, pp. 9570–9580.

Hu, Q., Zhang, Y., Zhang, R., et al., 2024. Beyond dehazing: learning intrinsic hazy robustness for aerial object detection. *IEEE Trans. Geosci. Remote Sens*. 62, 1–14.

Jeon, M., Seo, J., Min, J., 2024. Da-raw: domain adaptive object detection for real-world adverse weather conditions. *Proc. IEEE Int. Conf. Robot. Autom. (ICRA)*, pp. 2013–2020.

Li, C., Zhou, H., Liu, Y., et al., 2023a. Detection-friendly dehazing: object detection in real-world hazy scenes. *IEEE*

*Trans. Pattern Anal. Mach. Intell.* 45(7), 8284–8295.

Li, M., Xiong, G., Ye, P., et al., 2025a. Model with master-slave backbone and bifurcation fusion for UAV traffic object detection. *IEEE Trans. Instrum. Meas.* 74, 1–12

Li, Q., Zhang, Y., Fang, L., et al., 2025b. DREB-net: dual-stream restoration embedding blur-feature fusion network for high-mobility UAV object detection. *IEEE Trans. Geosci. Remote Sens. Early Access.*

Li, X., Diao, W., Mao, Y., et al., 2023b. OGMN: Occlusion-guided multi-task network for object detection in UAV images. *ISPRS J. Photogramm. Remote Sens.* 199, 242–257.

Li, Y.J., Dai, X., Ma, C.Y., et al., 2022. Cross-domain adaptive teacher for object detection. In: *Proc. IEEE/CVF Conf. Comput. Vis. Pattern Recognit. (CVPR)*, pp. 7581–7590.

Lin, X., Niu, Y., Yu, X., et al., 2025. Paying more attention on backgrounds: Background-centric attention for UAV detection. *Neural Netw.* 185, 107182.

Liu, S., Li, F., Zhang, H., et al., 2022a. DAB-DETR: Dynamic anchor boxes are better queries for DETR. *arXiv:2201.12329*.

Liu, W., Ren, G., Yu, R., et al., 2022b. Image-adaptive YOLO for object detection in adverse weather conditions. *Proc. AAAI Conf. Artif. Intell.*, pp. 1792–1800.

Ma, Y., Chai, L., Jin, L., et al., 2024. Hierarchical alignment network for domain adaptive object detection in aerial images. *ISPRS J. Photogramm. Remote Sens.* 208, 39–52.

Qin, Q., Chang, K., Huang, M., et al., 2022. Denet: detection-driven enhancement network for object detection under adverse weather conditions. *Proc. Asian Conf. Comput. Vis. (ACCV)*, pp. 2813–2829.

Ren, S., He, K., Girshick, R., Sun, J., 2015. Faster r-cnn: Towards realtime object detection with region proposal networks. Advances in neural information processing systems 28.

Sun, P., Zhang, R., Jiang, Y., et al., 2023. Sparse R-CNN: An end-to-end framework for object detection. *IEEE Trans. Pattern Anal. Mach. Intell.* 45(12), 15650–15664.

Tian, Z., Shen, C., Chen, H., He, T., 2020. Fcos: A simple and strong anchor-free object detector. IEEE Transactions on Pattern Analysis and Machine Intelligence.

Van der Maaten, L., & Hinton, G. (2008). Visualizing data using t-SNE. *Journal of Machine Learning Research*, *9*(11), 2579–2605.

Wang, G., Zhang, X., Peng, Z., et al., 2023. MOL: towards accurate weakly supervised remote sensing object detection via multi-view nOisy learning. *ISPRS J. Photogramm. Remote Sens.* 196, 457–470.

Wang, H., Liu, J., Zhao, J., et al., 2025. Precision and speed: LSOD-YOLO for lightweight small object detection. *Expert Syst. Appl.* 269, 126440.

Wang, J., Chen, Y., Zheng, Z., et al., 2024b. CrossKD: cross-head knowledge distillation for object detection. *Proc. IEEE/CVF Conf. Comput. Vis. Pattern Recognit. (CVPR)*, pp. 16520–16530.

Wang, K., Fu, X., Ge, C., et al., 2024a. Towards generalized UAV object detection: a novel perspective from frequency domain disentanglement. *Int. J. Comput. Vis.* 132(11), pp. 5410–5438.

Wang, R., Lin, C., Li, Y., 2025. RPLFDet: A lightweight small object detection network for UAV aerial images with rational preservation of low-level features. *IEEE Trans. Instrum. Meas.* Early Access.

Wang, W., Cai, Y., Wang, T., 2025. SRODET: Semi-supervised remote sensing object detection with dynamic pseudo-labeling. *IEEE Geosci. Remote Sens. Lett.* Early Access.

Wang, W., Cao, Y., Zhang, J., et al., 2021. Exploring sequence feature alignment for domain adaptive detection transformers. *Proc. 29th ACM Int. Conf. Multimedia*, pp. 1730–1738.

Yan, Z., Chen, C., Wu, S., et al., 2025. RF-DET: refocusing on the small-scale objects using aggregated context for accurate power transmitting components detection on UAV oblique imagery. *ISPRS J. Photogramm. Remote Sens.* 220, 692–711.

Zhang, B., Wang, Z., Du, B., 2024. Boosting semi-supervised object detection in remote sensing images with active teaching. *IEEE Geosci. Remote Sens. Lett.* 21, 1–5.

Zhang, H., Chang, H., Ma, B., et al., 2020. Dynamic R-CNN: Towards high quality object detection via dynamic training. In: *Proc. Eur. Conf. Comput. Vis. (ECCV 2020)*, Glasgow, UK. *Lect. Notes Comput. Sci.* 12360, pp. 260–275

Zhang, H., Xiao, L., Cao, X., et al., 2022. Multiple adverse weather conditions adaptation for object detection via causal intervention. *IEEE Trans. Pattern Anal. Mach. Intell*. 46(3): 1742-1756.

Zhang, Y., Gao, G., Chen, Y., et al., 2025. ODD-YOLOv8: an algorithm for small object detection in UAV imagery. *J. Supercomput.* 81(1), 1–17.

Zhao, Z., Wei, S., Chen, Q., et al., 2023. Masked retraining teacher-student framework for domain adaptive object detection. *Proc. IEEE/CVF Int. Conf. Comput. Vis. (ICCV)*, pp. 19039–19049.

Zhong, F., Shen, W., Yu, H., et al., 2024. Dehazing & reasoning YOLO: prior knowledge-guided network for object detection in foggy weather. *Pattern Recognit.* 156, 110756.